\documentclass[11pt,paper=a4]{article}
\usepackage{emnlp2017}
\usepackage{times}
\usepackage{latexsym}
\usepackage[utf8]{inputenc}
\usepackage[english]{babel}
\usepackage{url}
\usepackage{latexsym}
\usepackage{todonotes}
\usepackage{mathrsfs} % mathcal
\usepackage{amsmath} % align
\usepackage{amssymb} % nexists
\usepackage{mathtools} % coloneqq
\usepackage{xparse}
\usepackage{enumitem} % compact itemize
\usepackage[normalem]{ulem} % silbentrennung bei underline
\usepackage{listings}
\usepackage{graphicx} % graphics
\usepackage{hyphenat}
\usepackage{multirow}

\emnlpfinalcopy

\def\emnlppaperid{1}

\newcommand{\ignore}[1]{}
\newcommand{\cev}[1]

\title{Analysing Errors of Open Information Extraction Systems}

\author{Rudolf Schneider,  \bf{Tom Oberhauser},   \bf{Tobias Klatt}, \\ \bf{Felix A. Gers} \and \bf{Alexander Löser} \\
	 \{rudolf.schneider, tom.oberhauser, tobias.klatt, gers, aloeser\}@beuth-hochschule.de\\
	 Beuth University of Applied Sciences Berlin \\ Luxemburger Str. 10,
	 13353 Berlin, Germany}

\begin{document}\maketitle

\begin{abstract}
We report results on benchmarking Open Information Extraction (OIE) systems using RelVis, a toolkit for benchmarking Open Information Extraction systems.
Our comprehensive benchmark contains three data sets from the news domain and one data set from Wikipedia with overall 4522 labeled sentences and 11243 binary or n-ary OIE relations.
In our analysis on these data sets we compared the performance of four popular OIE systems, ClausIE, OpenIE 4.2, Stanford OpenIE and PredPatt.
In addition, we evaluated the impact of five common error classes on a subset of 749 n-ary tuples.
From our deep analysis we unreveal important research directions for a next generation of OIE systems. 
\end{abstract}

\section{Introduction}
Open Information Extraction (OIE) is an important intermediate step of the nlp stack for many text mining tasks, such as summarization, relation extraction, knowledge base construction and question answering \cite{mausam_open_2016, 2015stanovsky, khot_answering_2017}.
OIE systems are designed for extracting n-ary tuples from diverse and large amounts of text, without being restricted to a fixed schema or domain.
These tuples consist of one predicate and n arguments e.g: \textit{flew}(Obama; from Berlin; to New York).

Users often desire to select a suitable OIE system for their specific application domain.
Making the right choice is not an easy task.
Unfortunately, there is surprisingly little work on evaluating and comparing results among different OIE systems.
Worse, most OIE methods utilize proprietary and unpublished data sets.
In most cases users can only rely on publications and need to download, compile and apply existing systems to their own data sets.

\paragraph{Contribution} Ideally, one could compare different OIE systems with a unified benchmarking suite.
As a result, the user could identify ``sweet spots'' of each system but also weaknesses for common error classes.
The benchmarking suite should feature a diverse set of gold annotations with several thousands of annotated sentences.
By exploring results and errors, the user can learn how to design the next generation of OIE systems or how to combine several systems into an ensemble.

Our contributions are: (1) We report results of a quantitative analysis on four commonly used OIE systems: \textsc{Stanford OpenIE (SIE)} \cite{angeli_leveraging_2015}, \textsc{OpenIE 4.2 (OIE)}\footnote{https://github.com/allenai/openie-standalone}, \textsc{ClausIE (CIE)} \cite{del_corro_clausie:_2013} or \textsc{PredPat (PP)} \cite{white_universal_2016}.
Which employ rule based as well as machine learning based methods on linguistic structures like dependency parses.
These were applied on \emph{4522} sentences and \emph{11243} n-ary gold standard tuples.
(2) We share in-depth insights on a qualitative error analysis of \emph{749} n-ary tuples in \emph{68} sentences from four gold standard data sets annotated by all four OIE systems.
(3) We provide an integrated benchmark for OIE systems consisting of three news data sets NYT-222, WEB-500 \cite{mesquita-schmidek-barbosa:2013:EMNLP}, PENN-100 \cite{xu_open_2013} and a large OIE benchmark from Newswire and Wikipedia \cite{Stanovsky2016EMNLP} combined in our evaluation tool \textit{RelVis}. Our benchmark tool will be provided to the community under an open source license.

The remainder of this paper is structured as follows:
First, Section \ref{sec:analysing-oie} gives detailed insights on methods used for qualitative and quantitative evaluation.
Section \ref{sec:demo} introduces our evaluation system in a demo walkthrough.
We report in Section \ref{sec:results} in-depth insights on our experiment results.
Finally, Section \ref{sec:conclusion} concludes with design recommendations for next generation OIE systems. 
\section{Analysing Open IE Systems}\label{sec:analysing-oie}
 We set up two experiments with four OIE systems \textsc{Stanford OpenIE} \cite{angeli_leveraging_2015}, \textsc{OpenIE 4.2}\footnote{https://github.com/allenai/openie-standalone }, \textsc{ClausIE} \cite{del_corro_clausie:_2013} and \textsc{PredPat} \cite{white_universal_2016} and four gold standard data sets.
 The qualitative analysis was done by two human judges, who classified errors in the output of the systems into six categories.
 Our qualitative analysis includes gold labeled data sets from previous evaluations, shown in Table \ref{tbl:datasets}.

\subsection{Data sets}
 Our evaluation process for Open Information Extraction systems should be convenient and comparable. To meet this goal, we deliver supplementary scripts to import commonly used  data sets with our evaluation system RelVis. The unified data model enables the user to perform quantitative comparisons and extensive analyses on widely used data sets.
We used in our experiments four data sets, see Table \ref{tbl:datasets}, of which two feature only binary relations with two arguments.
Data sets \emph{NYT-222} and \emph{OIE2016} also contain n-ary relations. These labeled data sets origin from \newcite{mesquita-schmidek-barbosa:2013:EMNLP} and \newcite{Stanovsky2016EMNLP}.
\begin{table}[]
	\renewcommand{\arraystretch}{0.95}
	\setlength{\tabcolsep}{0.36em}
	\begin{tabular}{ l | c | c | c | rl}
		Name & Type & Domain & Sent. & \# Tuple \\ \hline
		\hline
		NYT-222 & n-ary & News & 222 & 222\\ \hline
		WEB-500 & binary & Web/News & 500 & 461\\ \hline
		PENN-100 & binary & Mixed & 100 & 51\\ \hline
		OIE2016 & n-ary & Wiki & 3200 & 10359\\ \hline
	\end{tabular}
	\caption{Data sets in RelVis \label{tbl:datasets}}
\end{table}

\subsection{Measuring OIE Systems}

A naive way to match a tuple to a gold standard is an \emph{equal match}.
Enforcing equal matching of boundaries in text to a gold standard delivers exact results for computing precision.
However, this strategy penalizes other, potentially correct, boundary definitions beyond the gold standard.
Dealing with multiple OIE systems and their different annotation styles requires a less restrictive matching strategy.
A second strategy is a \emph{containment match} where an argument or predicate is considered correct, if it at least contains a gold standard annotation.
Hence, spans from the gold standard may be contained (fully) inside the spans of the annotation from the OIE system.
However, this strategy may label over-specific tuples as correct and may lead to a lower precision and penalizes binary systems on n-ary data sets. 

Therefore we introduce a \emph{relaxed containment strategy} which removes a penalty for wrong boundaries especially for over-specific extractions.
This strategy counts an extraction correct, even when the number of arguments doesn't match the gold standard.
For example, Stanford OIE, a system that only returns binary OIE tuples, performs well on \textit{NYT-nary (b)}, an n-ary data set and yields large parts of relatively short sentences as one argument.
With the relaxed matching strategy Stanford OIEs binary extractions are counted correct as long as they contain all gold standard arguments.

The approach of \newcite{mesquita-schmidek-barbosa:2013:EMNLP} has simplified the task by replacing all entities in the test set with the words ``Europe'' and ``Asia''.
In our opinion this decision is contrary to the definition of OpenIE given by \newcite{banko_open_2007} which describes OIE as \textit{``domain-independent discovery of relations extracted from text and readily scales to the diversity and size of the Web corpus.''} and may hide or even cause problems in the analysed systems.

\paragraph{Measurements.}
In our quantitative evaluation we calculate precision, recall and $F_{2}$ measure at sentence level.
Following \newcite{pink_analysing_2014}, we choose $F_{2}$ instead of $F_1$ measure because it gives the recall a larger impact.
The basic intuition is that a high recall of an OIE system is critical to the performance of any downstream application that can apply additional filters.

\subsection{Common Error Classes.} \label{error_categories}
Authors of OIE systems distinguish among six major  error classes, Table \ref{tbl:qualitative-eval} reports errors for our four OIE systems. In the following paragraphs we describe each error class in detail.

\paragraph{Wrong Boundaries.}
\newcite{banko_open_2007} describe this error as \emph{too large or too small boundaries for an argument or predicate of an OIE extraction}.
In each of the four OIE systems we observe wrong boundaries for at least one third of the results.
This indicates that OIE systems often fail in generalizing to unseen word distributions.
This might be caused by errors in used intermediate structures, such as dependency parses, or overestimation of boundaries.
Incorrect boundaries for relation arguments can prohibit fusing, linking or aggregating tuples for the same predicate.
As a consequence, an additional system needs to filter out incorrect boundaries which may cause a drastic recall loss.

A solution proposed in the literature is to `wait' until intermediate systems, such as dependency parser, POS tagger etc., provide an improved generalization. However, this may not always be the case for niche domains, such as medical text or text in enterprise scenarios, where often no labeled corpora for those intermediate systems exist.

Following \newcite{DBLP:conf/coling/ArnoldDL16, arnold_robust_2016} we suggest end-to-end architectures, such as TASTY, an end-to-end named entity recognition and named entity linking system. TASTY leverage stacked deep learning architectures and requires only a few hundred labeled annotations to reach high F-measures in various domains and languages.

\paragraph{Redundant Extraction.}
In absence of a schema, OIE systems output redundant extractions for the same sentence, such as for the same subject-predicate structure.
For example, in the sentence \textit{``Additionally, we included some other relevant results from the 2005 survey in Antwerp.''} SIE yields two times the tuple (we, included, other relevant results).
These OIE systems are tuned towards high recall and leave the decision to filter out redundant tuples to a downstream application \cite{del_corro_clausie:_2013}.
The OIE system SIE which returns in extreme cases up to 140 tuples for the same sentence.
Our results indicate that this error class has been resolved to a large extent in most systems by filtering and aggregating results from multiple similar extraction rules.

\paragraph{Uninformative Extraction.}
Following \newcite{fader_identifying_2011}, \emph{uninformative extractions are extractions that omit critical information}.
This type of error is caused by improper handling of relation phrases that are expressed by a combination of a verb with a noun, such as light verb constructions (LVCs). Adding syntactic and lexical constraints may solve this problem to certain extent.

\paragraph{Missing Extraction - False Negatives.}
This class describes relations which were not found by a particular system.
According to \newcite{fader_identifying_2011}, missing extractions are often caused by argument-finding heuristics choosing the wrong arguments, or failing to extract all possible arguments. One example is the case of coordinating conjunctions.
CIE and OIE can spot certain cases of coordinating conjunctions and do miss fewer tuples. Other sources of this error are lexical constraints filtering out a valid relation phrase, another source are errors in dependency parsing.
Overall, we observe a trade-off among OIE systems between utilizing lexical constraints for filtering out uninformative tuples and thereby creating false negatives. Our results indicate that system OIE handles this trade-off slightly better than other systems.

\paragraph{Wrong Extraction.} \newcite{Stanovsky2016EMNLP} consider a tuple as correct as long as it shares a specified threshold of characters with a gold annotation. However, this policy may lead to emitting large parts of a sentence as one argument and poses additional computation effort to a downstream application. We focus on sentence-level correctness \cite{mesquita-schmidek-barbosa:2013:EMNLP, angeli_leveraging_2015} and define a tuple as correct if the following conditions are met:
\begin{enumerate}
	\item The selected matching strategy yields a match for the predicate.
	\item The number of arguments aligns with the gold standard.
	\item The selected matching strategy yields a match for all arguments.
\end{enumerate}
This error class is critical since it is not possible to recover from a error of this class and it emits a wrong signal which might trigger additional errors in downstream tasks.

\paragraph{Out of Scope.}
We observe in Table \ref{tbl:qualitative-eval} that our OIE systems yield more correct extractions as recognized by authors of gold data sets.
For these additional annotations, we introduce an \textit{out of scope} category.
This label does not indicate an error, but it helps us from distinguishing errors of gold labels and additional annotations of a particular OIE system that are not present in the gold standard.
Our two judges marked an annotation, in the qualitative evaluation, as out of scope if it is valid and provides an information gain.
If marked as out of scope, no other error category is applied to the extraction.

\section{The RelVis System}
\label{sec:demo}
In the following Section, we guide through our OIE benchmark system which was used to perform the quantitative and qualitative analysis. We show how the system can support a user in such sophisticated evaluation processes.

\paragraph{Startup.} At system initialisation, RelVis reads gold-annotations.
Next, the system stores extraction and gold annotations in a RDBMS from which a web based front end visualizes text data and annotations.

\begin{figure*}[]
	\centering
	\includegraphics[width=1\textwidth]{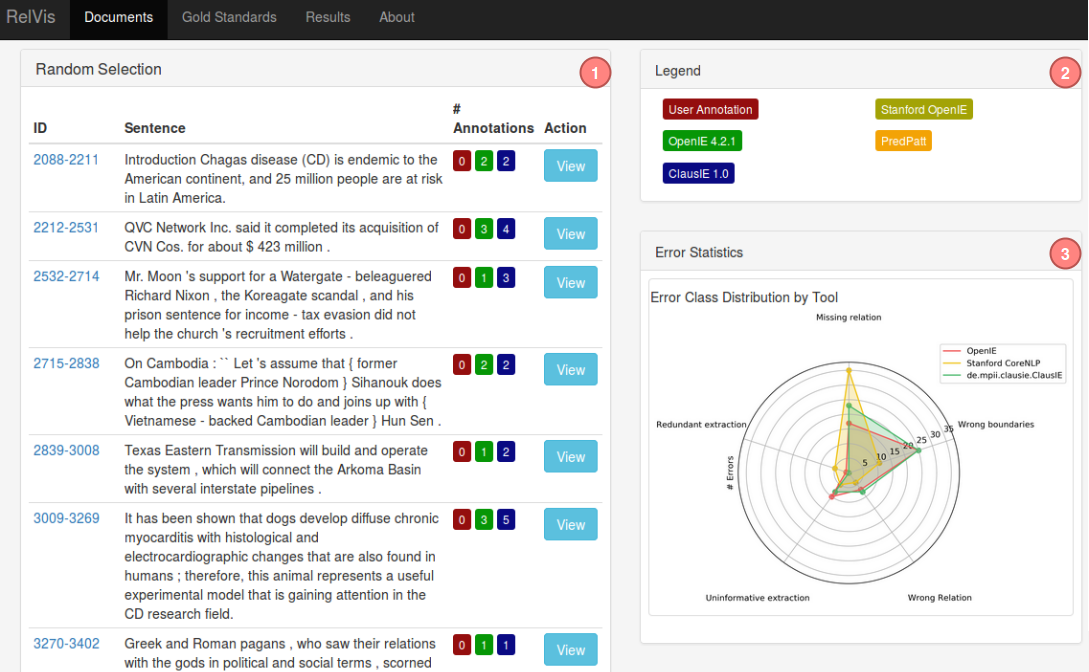}
	\caption{Screenshot of the sentence selection view of RelVis.
(1) For each sentence in the document we show text and number of extractions by system.
(2) The ``Legend panel'' denotes various OIE systems with different colours.
(3) The lower right hand side shows visualizations of error evaluation statistics. \label{fig:system-screenshot}}
\end{figure*}

\paragraph{Dashboards for exploring annotations.} Now, the user can start exploring results and understanding the behaviour of each system.
Figure \ref{fig:system-screenshot} visualizes in a dashboard example sentences,  precision, recall and $F_{2}$ scores for each OIE system and for each error class.

Please note, RelVis plots error distributions as a Kiviat diagram and draws bar charts for comparing error class impacts for each OIE system.
In addition, the user can export results as tables and CSV files from the database, as shown in Table \ref{tbl:qualitative-eval} and Table \ref{tbl:quantitative-result}.

\begin{figure*}[]
	\centering
	\includegraphics[width=1\textwidth]{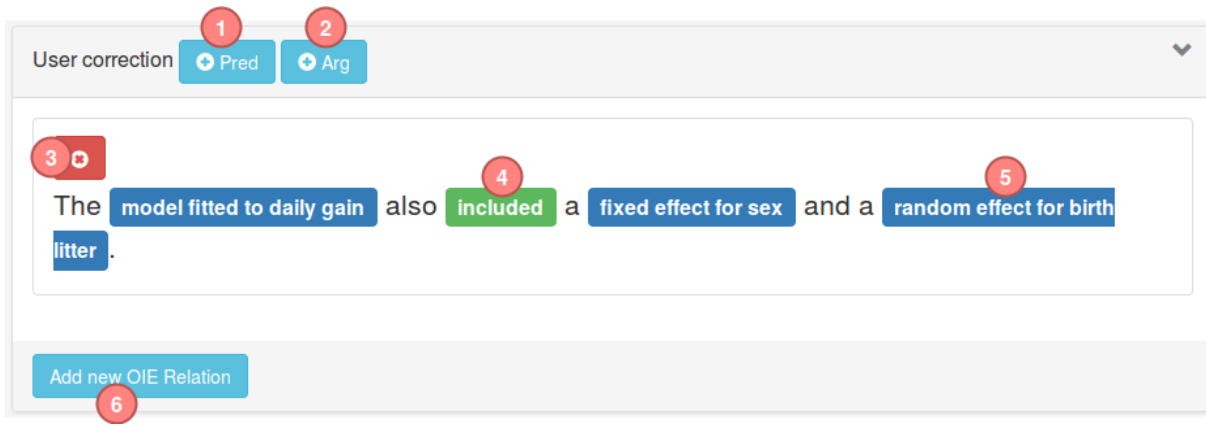}
	\caption{Visualization of OIE annotations. The predicate is marked green (4) and arguments blue (5). Buttons for adding a predicate (1) or argument (2) are on top. The button for adding another OIE Annotation (6) is on the bottom. In the top left corner is the delete annotation button (3). \label{fig:annotation-visualization}}
\end{figure*}

\paragraph{Understanding and adding a single annotation.} RelVis visualizes OIE extractions on sentence level.
Figure \ref{fig:annotation-visualization} shows how the dashboard visualizes example sentences.
For each hit by a system, the user can drill down into a single sentence and can understand extraction predicates or arguments.
\begin{figure}[t]
	\centering
	\includegraphics[width=0.5\textwidth]{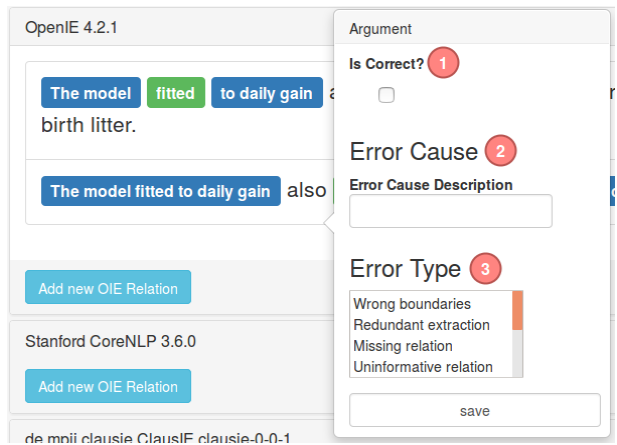}
	\caption{Interface for specifying the correctness (1), comment on error cause (2) and error class (3) of an OIE extraction. \label{fig:manual-evaluation}}
\end{figure}
Next, she can dive down into  correct or incorrect annotations, can add labels for error classes for incorrect annotations or may leave a comment, see also Figure \ref{fig:manual-evaluation}.
We permit the user to apply multiple error classes to each part of an annotation, such as a predicate or argument.
Next, she can focus on a sentence of interest and can compare extractions between different OIE systems.

We permit the user to update or add new annotations with a BRAT style functionality \cite{stenetorp_brat:_2012}, optimized for n-ary OIE relations.

Figure \ref{fig:annotation-visualization} shows a screenshot to illustrate the process.
The user selects a sentence to annotate and starts with the first annotation by clicking on the ``Add new OIE Relation'' button (6).
Next, she marks the predicate and arguments in the sentence for her first annotation by selecting them with the cursor and interacting with Button (1) and (2). The system indicates predicates (4) in green and arguments (5) in blue colour. 
\section{Experiment Results}\label{sec:results}

\begin{table*}[!ht]
	\centering
	\renewcommand{\arraystretch}{0.94}
	\setlength{\tabcolsep}{0.31em}
	\begin{tabular}{l|c|c|c|c|c|c|c|c|c|c|c|c}
		\textbf{Dataset} & \multicolumn{3}{c|}{\textbf{ClausIE  (\%)}} & \multicolumn{3}{c|}{\textbf{OpenIE 4.2  (\%)}} &
		\multicolumn{3}{c|}{\textbf{Stanford OIE  (\%)}} &
		\multicolumn{3}{c}{\textbf{PredPatt  (\%)}} \\
		& P & R & $F_{2}$ & P & R & $F_{2}$ & P & R & $F_{2}$ & P & R & $F_{2}$ \\
		\hline
		\hline
		PENN-100 (a) & 4.00 & 21.15 & 11.39 & 12.41 & 36.54 & 26.31 & \textbf{14.85} & \textbf{57.69} & \textbf{36.58} & 6.83 & 42.30 & 20.75 \\  \hline
		PENN-100 (b) & 4.00 & 21.15 & 11.39 & 13.07 & 38.46 & 27.70 & \textbf{14.85} & \textbf{57.69} & \textbf{36.59} & 7.76 & 48.08 & 23.58 \\  \hline \hline
		WEB-500 (a) & \textbf{16.33} & \textbf{46.70} & \textbf{34.03} & 12.83 & 19.62 & 17.74 & 13.65 & 40.72 & 29.16 & 5.18 & 13.43 & 10.19 \\  \hline
		WEB-500 (b) & \textbf{16.33} & \textbf{46.70} & \textbf{34.03} & 13.39 & 20.47 & 18.51 & 13.65 & 40.72 & 29.16 & 6.09 & 15.78 & 11.97 \\  \hline \hline
		NYT-222 (a) & 1.64 & 5.85 & 3.87 & \textbf{2.86} & 7.66 & 5.73 & 0 & 0 & 0 & 2.22 & \textbf{13.51} & \textbf{6.71} \\  \hline
		NYT-222 (b) & 4.69 & 16.67 & 11.03 & 11.28 & 30.18 & 22.60 & \textbf{13.37} & \textbf{73.87} & \textbf{38.77} & 8.47 & 51.35 & 25.51 \\  \hline \hline
		OIE2016 (a) & 14.81 & 13.67 & 13.89 & \textbf{24.85} & \textbf{18.69} & \textbf{19.67} & 0.80 & 1.49 & 1.27 & 7.26 & 12.39 & 10.86 \\  \hline
		OIE2016 (b) & 20.38 & 18.81 & 19.10 & \textbf{39.58} & \textbf{29.76} & \textbf{31.31} & 3.83 & 7.10 & 6.07 & 13.52 & 23.09 & 20.23 \\  \hline \hline
	\end{tabular}
	\caption{Quantitative Evaluation.
		The (b) variant are results with relaxed containment match strategy and (a) are those with the strict containment strategy.
		\label{tbl:quantitative-result}
	}
\end{table*}

\begin{table*}[t]
	\centering
	\setlength{\tabcolsep}{0.32em}
	\renewcommand{\arraystretch}{0.93}
	\begin{tabular}{l|c|c|c|c|c|c|c|c|c|c|c|c|c|c|c|c}
		\textbf{Dataset} & \multicolumn{4}{c|}{\textbf{NYT-222} (n-ary)} & \multicolumn{4}{c|}{\textbf{OIE2016} (n-ary)} & \multicolumn{4}{c|}{\textbf{PENN-100} (binary)} & \multicolumn{4}{c}{\textbf{WEB-500} (binary)} \\
		\# Relations & \multicolumn{4}{c|}{17} & \multicolumn{4}{c|}{29} & \multicolumn{4}{c|}{17} & \multicolumn{4}{c}{17} \\
		& CIE & OIE & PP & SIE & CIE & OIE & PP & SIE & CIE & OIE & PP & SIE & CIE & OIE & PP & SIE \\ \hline
		\# Predicted & 42 & 35 & 68 & 74 & 28 & 30 & 57 & 91 & 63 & 34 & 61 & 49 & 33 & 22 & 24 & 38 \\ \hline
		\# Correct & 2 & 1 & \textbf{6} & 0 & 8 & \textbf{12} & 6 & 5 & 4 & 8 & 10 & \textbf{11} & 5 & 4 & 3 & \textbf{10} \\ \hline \hline
		\# Redundant & 0 & 0 & 0 & \textbf{5} & 0 & 0 & 0 & \textbf{18} & 1 & 0 & 0 & \textbf{4} & \textbf{2} & 0 & 0 & 0 \\ \hline
		\# Uninformative & 4 & 2 & \textbf{8} & 0 & 2 & 0 & \textbf{6} & 1 & \textbf{9} & 3 & \textbf{9} & 4 & 0 & 0 & 0 & \textbf{3} \\ \hline
		\# Boundaries & 11 & 17 & 18 & \textbf{39} & 11 & 11 & 21 & \textbf{69} & \textbf{14} & 5 & 9 & \textbf{14} & 8 & \textbf{9} & \textbf{9} & \textbf{9} \\ \hline
		\# Wrong & 2 & 1 & 3 & \textbf{5} & 1 & 1 & \textbf{6} & 3 & 3 & 1 & \textbf{10} & 4 & 1 & \textbf{2} & \textbf{2} & \textbf{2} \\ \hline
		\# Out of Scope & 24 & 17 & \textbf{34} & 30 & 7 & 6 & \textbf{21} & 13 & \textbf{33} & 17 & 31 & 18 & \textbf{19} & 8 & 12 & 14 \\ \hline \hline
		\# Missed & 4 & 1 & \textbf{5} & \textbf{5} & 8 & 4 & 7 & \textbf{12} & \textbf{14} & 6 & 6 & 7 & 8 & 3 & \textbf{11} & 6 \\ \hline
	\end{tabular}
	\caption{Occurrences of extraction errors found in the qualitative analysis of four OIE systems on 17 sentences drawn from four gold standard datasets.
	749 predicted extractions were evaluated in total.
	Note: multiple errors per predicted extraction are possible and that number of missed extractions is naturally not contained in \# Predicted.}
	\label{tbl:qualitative-eval}
\end{table*}

\begin{figure}[h]
	\centering
	\includegraphics[width=0.48\textwidth]{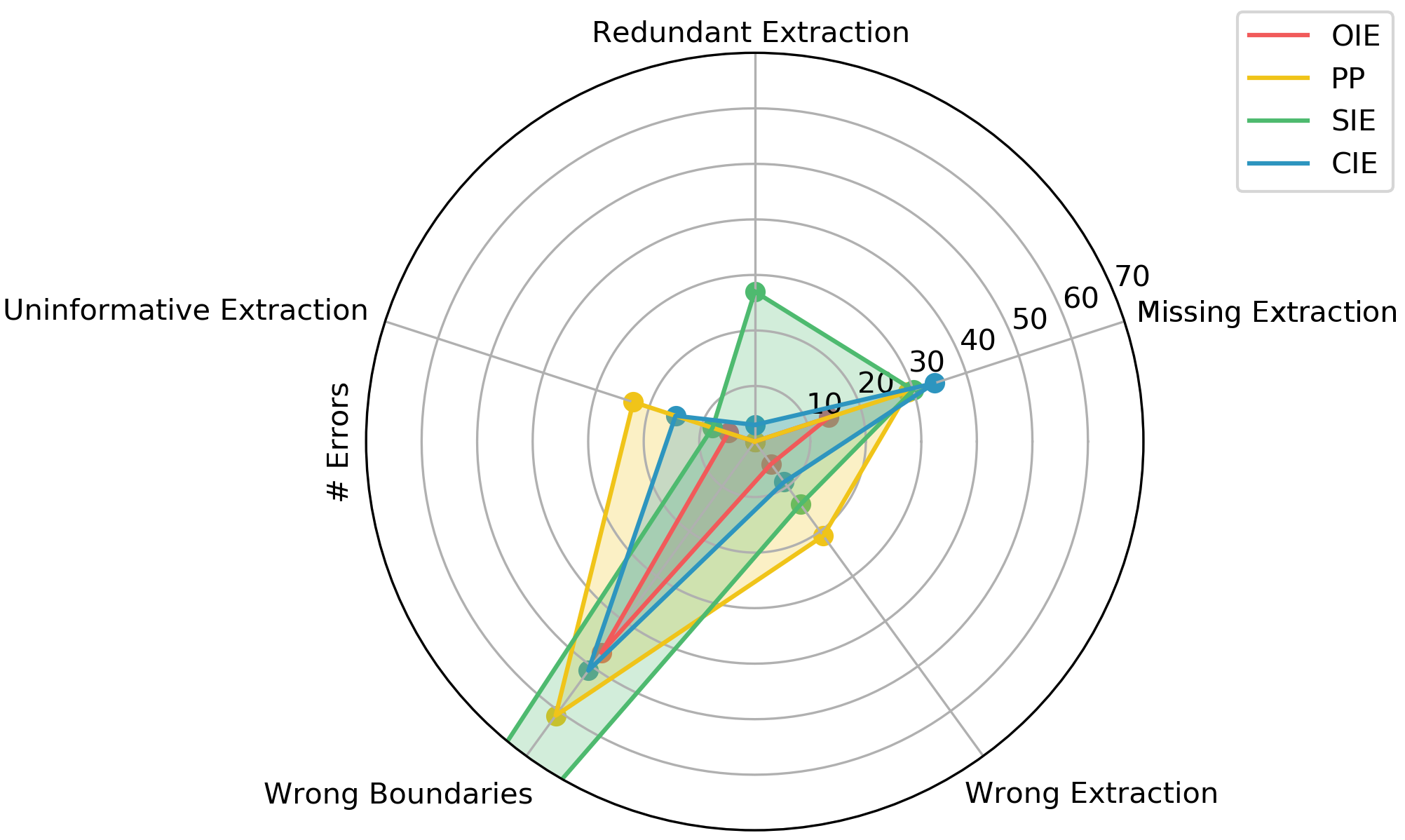}
	\caption{Error occurrence of all OIE systems on 68 sentences of four data sets.
		Error categories described in \ref{error_categorusies} are plotted along five axes.
		The system with the smallest covered area makes the least errors.
	We crop the diagram at 70 occurrences for easier interpretation.
	SIE hits 131 times in total the Wrong Boundaries category.
		\label{fig:number_of_errors}}
\end{figure}
In a \textbf{quantitative evaluation} we report precision, recall and $F_2$ scores on all four data sets.
Table \ref{tbl:quantitative-result} reports overall results for four OIE systems on all four data sets, with the limitation that only a subset of OIE2016, containing 1768 sentences, was available to us.
We conduct our experiments with an exact (a) and relaxed (b) containment match strategy. 

For the \textbf{qualitative evaluation} we execute four OIE systems on 17 sentences of each data set.
This resulted in 749 predicted extractions which we evaluate and classify into error categories by two human judges, as shown in Table \ref{tbl:qualitative-eval}.
Additionally, Figure \ref{fig:number_of_errors} gives an overview of the general performance of all tools over all data sets.
We apply a strict containment match strategy in this evaluation.
Observing that multiple errors can happen to a single extraction, we assign in these cases more than one error category.

Note, for both experiments, we configure system CIE to binary extraction mode for binary data sets and otherwise in n-ary mode.

\subsection{General Findings}
We observe no clear overall winner: Each OIE system works best on a particular data set, and no OIE system significantly outperforms on two or more data sets. 

\paragraph{Boundary Errors.}
We observe that an OIE system causes boundary errors often by over- or under-specific argument spans.
In more rare cases the source for this error are predicate spans.
Both, argument and predicate related errors can be caused by wrong intermediate structures in a particular OIE system.
Another source of the problem could be the argument candidate generation, which overestimates the size of an argument span, so that it envelops multiple distinct arguments.
Further causes for a boundary error are different annotation styles, which appear among systems as well as among gold standard data sets.

As one possible source for the overall bad results on the NYT-222 dataset, we pinpoint the differing styles of conjunction extraction.
Consider a gold standard which expects a single extraction with multiple arguments for the sentence: ``\textit{DENVER BRONCOS signed LB Kenny Jackson, DT Garrett Johnson and CB Sam Young.}'' like e.g. \textit{signed(DENVER BRONCOS; Kenny Jackson; Garrett Johnson; Sam Young)}.
Systems CIE and OIE yield persons and their positions as one large argument in a binary relation:
\textit{signed(DENVER BRONCOS; LB Kenny Jackson, DT Garrett Johnson and CB Sam Young.)}. On the contrary, System PP implements another style extracting every person of the sample sentence in an own binary relation.

SIE, a binary extraction system, performs surprisingly well on this data set with the relaxed containment match strategy and on \textit{NYT-222 (b)}.
With a strict containment match strategy, NYT-222 (a), the system was not able to find a correct extraction, because the data set does not contain binary relations.
Using a relaxed containment match strategy, system SIE outperforms all other systems, by extracting large, over-specific arguments.
This shifts additional effort for further processing towards downstream applications.
This shows the importance of taking boundaries into account in an evaluation.
However, system SIE fails on the extraction of OIE2016, which contains more complex sentences, including numerical values and multiple gold annotations in comparison to NYT-222.

\paragraph{Missed Extractions.}
Noisy text, wrong intermediate structures and different annotation styles among gold data sets often trigger this error.
We report a significant drop in recall for all systems on the WEB-500 dataset compared to PENN-100, except for CIE, see Table \ref{tbl:quantitative-result}, even though both data sets show a similar annotation style.
However, the WEB-500 data set is quite noisy and contains HTML-character encodings, unfinished sentences or headlines with special characters.
Those artifacts cause errors in intermediate structures, like dependency parses or POS tags, which causes the systems to fail.
In particular, the n-ary systems OIE or PP do not seem to be robust to such noisy data.

Another source for missed relations is a mismatch between annotation styles.
For example, system CIE shows a different style as the gold annotation in PENN-100, NYT-222 and WEB-500 data sets.
A closer inspection reveals that CIE's verb centric extraction behaviour handles nominal or adjectival triggered relations \cite{peng_generalizable_2014} in a different style as the gold standard data set.
Its design triggers inserting an artificial predicate \cite{del_corro_clausie:_2013} which can cause many missed annotations in our evaluation.
For example, consider the following  sentence: ``\textit{At least one potential GEC partner, Matra, insists it isn't interested in Ferranti.}'' System CIE extracts the tuple: \textit{is(one potential GEC partner; Matra)}, but the style of the gold standard expects: \textit{partner(GEC; Matra)}.
We explain the increase of all scores of system CIE by the larger number of gold annotations, compared to PENN-100, which does not interfere with the annotation style of system CIE.

\paragraph{Wrong and uninformative unary extractions.}
Wrong extraction errors are in many cases complex and caused by other errors.
For example, a boundary error often leads to missing a important information like a negation.
Furthermore, we observe problems in the predicate candidate selection process for unary extractions which leads to wrong extractions.

Uninformative extractions are mostly yielded by systems CIE and PP. In many cases, these errors are triggered in possessive relations without resolved co-references or relations with adjectival triggers, e.g. \textit{first(world war)}. To overcome these problems, we suggest to improve filtering for uninformative unary relations, supply additional checks for missed negations or important arguments and integrate a co-reference resolution components into next generation OIE systems.

\paragraph{Redundant Extractions}
Redundant extractions exclusively occur in systems SIE and CIE\footnote{in binary extraction mode}. 

\subsection{Data set Specific Findings}

\paragraph{OIE systems are still designed towards binary tuples.}
Very first OIE systems had been designed towards emitting binary OIE tuples. Therefore, we observe that all systems  achieve a better recall score on the binary data sets when the strict containment strategy is used.
This is caused by larger number of possible errors in an n-ary task. Additionally, inconsistent extraction styles for n-ary relations in both, systems and gold standards,  cause errors. 

\paragraph{Out of Scope.}

The PENN-100 data set supplies for every sentence just one gold extraction.
In most cases it represents a non verbal triggered relation.
This leads to many out of scope extractions, because most of the systems perform well in extracting verbal triggered relations.
Each OIE system yields out of scope extractions in particular on the NYT-222 data set, which shows that the gold annotations in this data set do not cover capabilities of modern OIE systems.

Data set OIE2016 features the lowest number of out of scope extractions overall.
It provides multiple gold annotations per sentence and covers a wide variety of extractions, starting with unary up to 7-ary tuples.
System PP yields non verbal triggered unary extractions more often than other systems, which is the reason for its steady high number of out of scope extractions.

\section{Conclusion}\label{sec:conclusion}
To our best knowledge this is the first attempt of a comprehensive in-depth error analysis, containing quantitative and qualitative evaluations, of four OIE systems on four data sets.
In our future work we will publish our benchmark system RelVis, data sets and adapters under an open source licence for the general OIE community.\footnote{https://github.com/SchmaR/RelVis}

Because of the nature of the OIE task, we conclude that there is a lack in stringent annotation policies, which makes a comparative analysis but also the design of OIE system often difficult.
Moreover, each tested OIE system depends on syntactic taggers that often propagate errors towards the logic for extracting OIE tuples.
We also observe fewer errors among binary OIE tuples.
This indicates that current OIE systems have not reached an effective design yet for extracting higher order n-ary tuples.
Only system PP leverages well researched ideas from normal forms in data base theory in its design.

We suggest designers of next generation OIE systems to test their systems against various data sets, even data sets in idiosyncratic domains not included in this benchmark.
Moreover, next generation OIE systems should offer some convenient `knobs' for tuning it towards common downstream tasks, such as populating a knowledge base or extracting typed relations against a schema.

\section*{Acknowledgements}
Our work is funded by the German Federal Ministry of Economic Affairs and Energy (BMWi) under grant agreement 01MD16011E (Medical Allround-Care Service Solutions), grant agreement 01MD15010B (Smart Data Web) and by the European Union’s Horizon 2020 research and innovation programme under grant agreement No 732328 (FashionBrain).

\bibliography{emnlp2017}
\bibliographystyle{emnlp_natbib}

\end{document}